\documentclass[11pt, letterpaper]{article}

\usepackage[margin=1.2in]{geometry} 
\usepackage{amsmath, amssymb, amsfonts} 
\usepackage{graphicx} 
\usepackage{hyperref} 
\usepackage{natbib} 
\usepackage{booktabs} 
\usepackage{microtype} 
\usepackage{xcolor} 
\usepackage[T1]{fontenc}
\usepackage{lmodern}
\usepackage{microtype} 
\usepackage{placeins}
\usepackage{booktabs}
\usepackage{multirow}
\usepackage{graphicx}
\usepackage{subcaption} 
\usepackage{indentfirst}

\hypersetup{
    colorlinks=true,
    linkcolor=blue!60!black,
    citecolor=blue!60!black,
    urlcolor=blue!60!black
}

\title{\textbf{The Geometric Inductive Bias of Grokking:\\ Bypassing Phase Transitions via Architectural Topology}}
\author{
    \textbf{Alper YILDIRIM} \\
    Independent Researcher \\
    \texttt{yildirim.alper.dev@gmail.com}
}
\date{} 

\begin{document}

\maketitle

\begin{abstract}
Prior work on grokking---delayed generalization after memorization---has sought to shorten this delay through data augmentation or optimizer design. We take an architectural approach: rather than analyzing trained networks post-hoc, we modify architectural topology \emph{before} training to test whether specific degrees of freedom prolong the memorization phase. We identify two independent factors in standard Transformers---unbounded residual magnitude and data-dependent attention routing---and introduce targeted interventions for each. A fully bounded spherical topology, enforcing $L_2$ normalization throughout the residual stream, eliminates the memorization phase entirely on modular addition and multiplication over $\mathbb{Z}_p$: training and test accuracy rise together from initialization. Independently, replacing learned attention with uniform aggregation achieves the same effect, consistent with theoretical results showing that commutative operations require only a bag-of-tokens representation. However, the same spherical constraint \emph{fails completely} on non-commutative $S_5$ permutation composition, where successful solutions rely on discrete coset structures rather than continuous Fourier features. This contrast demonstrates that bypassing the generalization delay is possible---but strictly depends on alignment between architectural priors and task symmetry. Since even the simplest non-commutative task resists the same constraint that eliminates delay on cyclic arithmetic, these results suggest that no single geometric prior can universally accelerate learning across the heterogeneous structures present in general-purpose domains.
\end{abstract}

\vspace{0.5cm}


\section{Introduction}

Neural networks often exhibit complex and unintuitive learning dynamics. One of the most striking examples is grokking: a delayed phase transition in which a model achieves near-perfect training accuracy while test accuracy remains low, followed by a sudden transition to full generalization after prolonged optimization. Originally observed in small models trained on algorithmic tasks \citep{power2022grokking}, grokking has become a central phenomenon for studying the relationship between optimization, generalization, and implicit bias. Prior efforts to shorten this delay have operated at the level of data augmentation \citep{abramov2025grokking} or optimizer design \citep{lee2024grokfast}. In this work, we take a third approach: architectural intervention.

Mechanistic interpretability has shown that Transformers trained on modular arithmetic eventually implement structured representations based on discrete Fourier features \citep{nanda2023progress}. However, much of this work is post-hoc: researchers analyze models only after grokking has occurred, inferring mechanisms from frozen weights. We instead adopt a complementary interventional methodology. Rather than analyzing a trained model, we modify architectural structure \textit{before} training to test whether specific representational degrees of freedom influence grokking dynamics.

Our central hypothesis is that standard Transformer architectures possess degrees of freedom that exceed the minimal symmetry requirements of commutative, periodic tasks such as modular arithmetic. As demonstrated by \citet{zhong2023clock}, this excess capacity allows standard networks to converge on diverse, sometimes fragmented, piecewise solutions (the ``Pizza'' algorithm) rather than the structured, continuous Fourier solution (the ``Clock'' algorithm). We hypothesize that these additional degrees of freedom enable memorization-heavy solution pathways that delay the emergence of invariant representations \citep{zhong2023clock, lei2025construct, prieto2025grokking}. We isolate and study two independent architectural factors:

\paragraph{Path A: The Magnitude Degree of Freedom.}
In standard Transformers, information can be encoded in both the direction and magnitude of residual stream vectors. Prior work has shown that unconstrained magnitude growth can influence training dynamics in grokking settings \citep{prieto2025grokking}. To examine the role of this degree of freedom, we introduce a \textbf{Fully Bounded Spherical Topology} that enforces strict $L_2$ normalization throughout the residual stream and applies a normalized unembedding matrix with a fixed temperature scale. This removes the model's ability to encode information in vector norm and constrains logits to a bounded cosine geometry. Empirically, this modification qualitatively alters the learning dynamics on both modular addition and multiplication over $\mathbb{Z}_p$: rather than exhibiting the characteristic memorization plateau followed by delayed generalization, bounded models show training and test accuracy rising concurrently from initialization, with no separable memorization phase, while still using fourier circuits.
\paragraph{Path B: The Routing Degree of Freedom.}
Transformers also possess flexible, data-dependent attention routing through learned query-key interactions. However, recent theoretical results demonstrate that modular addition can be implemented using uniform token aggregation \citep{huang2025provable}. Motivated by this, we introduce a \textbf{Uniform Attention Ablation} that overrides learned query-key scores with a fixed uniform distribution, eliminating adaptive routing and reducing the attention mechanism to a Continuous Bag-of-Words (CBOW) aggregator. Despite removing routing flexibility, LayerNorm models with uniform attention achieve 100\% peak test accuracy across all seeds and bypass the grokking delay entirely.

\paragraph{Summary of Contributions.}
Through controlled architectural interventions, we show that constraining representational magnitude and removing data-dependent routing each independently eliminate the grokking delay on cyclic modular arithmetic (addition and multiplication over $\mathbb{Z}_p$). However, using non-commutative $S_5$ permutation composition as a negative control, we find that the same spherical constraint fails entirely: models cannot generalize on any seed within the training window, despite reaching perfect training accuracy. Prior interpretability work has shown that successful $S_5$ solutions rely on discrete coset structures \citep{stander2024grokking} rather than the continuous Fourier features that align with spherical geometry. This differential outcome demonstrates that bypassing the memorization phase depends on alignment between architectural priors and the task's intrinsic symmetry---it is not a generic effect of constraining representational capacity.

This finding carries a broader implication. Since even the simplest non-commutative algebraic task resists the same constraint that eliminates delay on cyclic arithmetic, it is unlikely that any single geometric prior can universally accelerate learning across the heterogeneous structures present in general-purpose domains such as natural language. Our results support using structural alignment not as a universal remedy, but as a diagnostic and interventional tool: a framework in which mechanistic analysis informs architectural design, and architectural design provides experimental tests of interpretability hypotheses.
\section{Related Work}

Prior work on grokking \citep{power2022grokking} spans three levels of intervention: data manipulation \citep{abramov2025grokking}, optimizer design \citep{lee2024grokfast}, and architectural structure. We contribute to the third.

\paragraph{Observational Foundations.}
Mechanistic analyses have shown that Transformers trained on modular addition construct Fourier-based representations mapping inputs to roots of unity on a complex circle \citep{nanda2023progress}. \citet{zhong2023clock} identified two qualitatively distinct solutions: a structured ``Clock'' algorithm using continuous Fourier features and a fragmented ``Pizza'' algorithm relying on piecewise memorization, demonstrating that unconstrained architectures can converge on either. For non-commutative tasks such as $S_5$ permutation composition, successful solutions instead rely on higher-dimensional irreducible representations and coset circuits \citep{chughtai2023neural, stander2024grokking}. Geometrically, the grokking delay has been linked to slow compression of the representation manifold \citep{zheng2024delays, lei2025construct, minegishi2026emergent}, while \citet{morwani2024feature} showed theoretically that the emergence of structured circuits is driven by implicit margin maximization under cross-entropy loss---the grokking delay corresponding to the time required for this bias to sculpt optimal geometry from an unstructured parameter space. \citet{musat2026geometry} formally proved that post-memorization dynamics approximate norm minimization on the zero-loss manifold under weight decay.

\paragraph{Architectural Inductive Biases.}
Several works have demonstrated that embedding periodic structure into network components accelerates learning on algorithmic tasks. \citet{huang2025provable} proved a formal expressivity gap: sinusoidal activations require only constant width for modular addition, whereas ReLU networks require width scaling with the modulus. Crucially, they also established that modular addition can be perfectly realized using a uniform sum of input tokens, motivating our Intervention~B. Other approaches include sinusoidal initialization \citep{fernandez2025sinusoidal}, Fourier-constrained output heads \citep{gillman2025fourier}, and the observation that pre-trained LLMs natively develop Fourier features for arithmetic while models trained from scratch do not \citep{zhou2024pretrained}. \citet{prieto2025grokking} showed that unconstrained logit growth leads to Softmax Collapse, motivating our bounded unembedding design. In concurrent work, \citet{singh2026explaining} found that restricting LayerNorm placement can accelerate generalization by limiting reliance on activation scale.

Our approach differs from these works in targeting the residual stream topology directly. Rather than altering activations, initialization, output heads, or normalization placement, we impose a hard $L_2$ spherical constraint on the residual stream itself, testing whether eliminating the radial degree of freedom can bypass the memorization phase entirely. The optimizer-level approach of \citet{lee2024grokfast}---which accelerates grokking up to $50\times$ via gradient low-pass filtering---is complementary but operates within the memorize-then-generalize paradigm; our architectural interventions eliminate the memorization phase rather than shortening it.

Most directly related to our Intervention A, Loshchilov et al.\ \cite{loshchilov2025ngptnormalized} proposed nGPT, a normalized Transformer that constrains all vectors to the unit hypersphere, achieving 4--20$\times$ training speedups on language modeling. Our spherical constraint is deliberately more minimal, as our goal is not faster pretraining but isolating how the radial degree of freedom influences the memorization-to-generalization transition.

\textbf{Concurrent Work.}\quad In concurrent independent work, Truong et al.~\cite{truong2026norm} derive tight
bounds on the grokking delay, establishing that $T_{\mathrm{grok}} - T_{\mathrm{mem}} = \Theta\bigl(\gamma_{\mathrm{eff}}^{-1} \log \tfrac{\|\theta_{\mathrm{mem}}\|^{2}}{\|\theta_{\mathrm{post}}\|^{2}}\bigr)$, where $\gamma_{\mathrm{eff}}$
is the effective contraction rate under weight decay. By structurally removing magnitude as a representational degree of freedom, our spherical constraint prevents the formation of high-norm memorization interpolants. This collapses the parameter norm ratio $\|\theta_{\mathrm{mem}}\|^2 / \|\theta_{\mathrm{post}}\|^2$ toward unity and predicts zero delay---consistent
with our empirical observation that bounded models generalize without a memorization phase. We discuss the implications for task-specific alignment in Section 5.1.

\section{Formal Problem Setting and Methodology}
\label{sec:methodology}

To investigate the structural drivers of delayed generalization (grokking), we build upon the mechanistic interpretability framework established by \citet{nanda2023progress}. Their work demonstrated that transformers trained on modular addition construct continuous Fourier-based representations. In particular, learned attention heads organize token interactions such that downstream MLP layers compute trigonometric identities consistent with circular phase structure \cite{zhou2024pretrained}.

Standard transformer architectures, however, contain representational degrees of freedom that exceed the minimal mathematical requirements of commutative, periodic tasks. We identify two distinct excess degrees of freedom: unbounded vector magnitude in the residual stream, and asymmetric, data-dependent attention routing. We hypothesize that these additional degrees of freedom permit solution pathways that do not respect task symmetry, enabling symmetry-misaligned solution pathways that delay convergence to the structured Fourier circuit. \cite{zheng2024delays, lei2025construct}.

To evaluate this hypothesis, we introduce two independent structural interventions to isolate each degree of freedom:
\begin{enumerate}
    \item \textbf{The Spherical Residual Stream (\S\ref{sec:spherical}):} Constrains magnitude growth to inject a geometric inductive bias aligned with Fourier representations.
    \item \textbf{Uniform Attention Ablation (\S\ref{sec:uniform_attention}):} Ablates data-dependent routing to test if complex attention is merely a memorization artifact, reducing aggregation to a theoretically optimal uniform sum.
\end{enumerate}

Finally, to distinguish task-specific geometric alignment from generic stabilization effects, we introduce symmetric group composition ($S_5$) as a negative control (\S\ref{sec:s5_task}).

\subsection{The Standard Residual Stream (Baselines)}
\label{sec:standard_baseline}

In standard transformer architectures, the residual stream $h$ serves as a cumulative communication channel across layers. In the absence of strict topological constraints, information may be encoded in both the direction (angle) and the magnitude of the state vector. For a given layer $l$, the unconstrained forward pass is:

\begin{align}
    h_{l}^{(mid)} &= h_{l-1} + \text{Attention}(h_{l-1}) \\
    h_{l} &= h_{l}^{(mid)} + \text{MLP}(h_{l}^{(mid)})
\end{align}

Modern implementations commonly include normalization mechanisms such as LayerNorm or RMSNorm to stabilize optimization. However, these methods do not impose a strict bound on residual magnitude. Although activations are normalized statistically, learnable affine parameters—particularly the scaling parameter $\gamma$—allow the network to rescale representations after normalization. As a result, magnitude remains an adjustable representational degree of freedom.

Recent theoretical and empirical work suggests that overparameterized models initially converge to high-frequency or memorization-based solutions before discovering structured representations \cite{gillman2025fourier, zheng2024delays}. Motivated by these findings, we hypothesize that retained magnitude flexibility in the residual stream provides a high-variance pathway for encoding training-specific noise, fitting the training data without aligning with the task’s underlying periodic structure.

\subsection{The Spherical Residual Stream and Fully Bounded Topology (Intervention A)}
\label{sec:spherical}

To examine the role of magnitude variation, we introduce the \textbf{Spherical Residual Stream}. We define a projection operator $\Pi_S$ that applies strict $L_2$ normalization across the feature dimension:

\[
\Pi_S(x)=\frac{x}{\max(\|x\|_2, \epsilon)}
\]

where $\epsilon=10^{-12}$ ensures numerical stability. This operator is applied to the residual stream prior to each sub-layer and immediately after each residual addition. A single transformer block is thus modified as:

\begin{align}
    h_{in} &= \Pi_S(h_{l-1}) \\
    h_{mid} &= \Pi_S(h_{in} + \text{Attention}(h_{in})) \\
    h_{l} &= \Pi_S(h_{mid} + \text{MLP}(h_{mid}))
\end{align}

This mechanism replaces standard LayerNorm, fixing the global residual norm to unity at each projection step. 

Crucially, this spherical constraint acts as a direct \textbf{geometric inductive bias} by eliminating radial degrees of freedom in the residual stream. In high-dimensional spaces, unconstrained magnitude allows representations to spread outward and partition space using norm-based separation. As demonstrated by \citet{zhong2023clock}, such expansive magnitude freedom enables the construction of highly fragmented, disjoint decision regions characteristic of the memorization-heavy ``Pizza'' algorithm.

By projecting activations onto a fixed-norm hypersphere, we remove magnitude as a representational axis, restricting the model to encode information purely through angular relationships. While the hypersphere remains $(d_{\text{model}}-1)$-dimensional, eliminating radial variability substantially constrains how the space can be partitioned. This restriction reduces the network’s ability to form piecewise magnitude-based memorization basins and instead biases learning toward continuous angular structure. 

In modular addition, this angular structure aligns naturally with 1D Fourier modes (the ``Clock'' algorithm). Empirically, we observe that enforcing this constraint from initialization dramatically shortens the delayed generalization phase, consistent with the hypothesis that removing magnitude-based representational freedom accelerates convergence toward structured solutions.

\paragraph{Bounded Output Topology.}
The spherical residual stream alone is insufficient: if the final unembedding matrix ($W_{\text{unembed}}$) remains unconstrained, training under cross-entropy loss can drive increasing logit magnitudes to artificially lower the loss. This behavior—known as Naïve Loss Minimization—leads to numerical instability and Softmax Collapse, which can severely delay or crash generalization \cite{prieto2025grokking}. Empirically, we found that all spherical models required unembedding normalization to train stably.

To stabilize the geometry and prevent Softmax Collapse, we apply $\Pi_S$ to the unembedding weights and compute logits via scaled cosine similarity:
\[
\text{Logits}=\tau \left( \Pi_S(h_{\text{final}})\,\Pi_S(W_{\text{unembed}}) \right)
\]

Since cosine similarity lies in $[-1,1]$, logit magnitudes are strictly bounded by the temperature parameter $\tau$ (set to $10.0$ in our experiments). By constraining both the internal residual geometry and the output logit scale, this \emph{fully bounded spherical topology} allows the network to stably lock into the Fourier solution without relying on weight decay to control magnitude growth.

\paragraph{Fourier Initialization.}
In addition to the standard random initialization, we evaluate a variant in which the first 10 dimensions of the token embedding matrix are deterministically initialized with cosine and sine values at five key frequencies, following the Fourier structure identified in prior mechanistic analyses \citep{nanda2023progress}. This provides the model with an explicit periodic prior from initialization. We refer to this variant as \textbf{Bounded Sphere + Fourier Init}.

\subsection{Uniform Attention Routing (Intervention B)}
\label{sec:uniform_attention}

In standard unconstrained transformers, attention mechanisms possess the representational capacity to learn complex, data-dependent query-key routing. However, recent theoretical proofs establish that modular addition ($a+b \pmod p$) can be perfectly realized by a network operating on a simple, uniform, unweighted sum of the input tokens (a ``bag-of-tokens'' vector) \citep{huang2025provable}. 

This implies that for strictly commutative operations, complex query-key routing is an unnecessary degree of freedom. We hypothesize that this excess flexibility permits the network to construct asymmetric representations that memorize specific $(a,b)$ token pairs, acting as an independent driver of delayed generalization.

To formally test whether learned routing is computationally required to solve the task, or if it merely traps the model in a memorization basin, we introduce a secondary, independent architectural intervention: the \textbf{Uniform Attention Ablation}. In this configuration, we completely ablate the data-dependent query-key routing by overriding the pre-softmax attention scores to zero:

\begin{equation}
    \text{Scores} = \frac{QK^T}{\sqrt{d_{\text{head}}}} \rightarrow \mathbf{0}
\end{equation}

Passing this zeroed matrix through the standard Softmax operator forces a perfectly uniform attention distribution across the sequence. For a sequence containing three tokens (e.g., $a$, $b$, and $=$), the attention weights become fixed at $[1/3,\,1/3,\,1/3]$. This intervention reduces the attention layer to a data-independent Continuous Bag-of-Words (CBOW) aggregator, structurally enforcing permutation invariance and perfectly matching the theoretical bag-of-tokens requirement \cite{huang2025provable}.

\subsection{Task 1: Cyclic Arithmetic over $\mathbb{Z}_p$ (Geometric Alignment)}
Prior mechanistic interpretability studies show that modular addition ($\mathbb{Z}_p$) is frequently implemented via discrete Fourier representations, mapping integer inputs to the roots of unity on a complex circle \cite{nanda2023progress}. In our experiments, we set $p=113$. We generate the exhaustive set of all $p \times p$ pairs, formatting the input sequences as \texttt{[a, b, equals\_token]}. To induce the delayed generalization characteristic of grokking, we train on a constrained fraction of the data, using a random $30\%$ split for training ($\sim 3,830$ samples) and the remaining $70\%$ for testing ($\sim 8,939$ samples):

\begin{equation}
E(x)_k = \exp\left(i \frac{2\pi k x}{p}\right) = \cos\left(\frac{2\pi k x}{p}\right) + i \sin\left(\frac{2\pi k x}{p}\right)
\end{equation}

In standard unconstrained networks, the model must simultaneously learn to align phases, suppress magnitude variations, and route tokens appropriately to compute these trigonometric interactions. We expect that by isolating and structurally removing the degrees of freedom that allow for non-Fourier memorization—either by mathematically bounding the geometry (Intervention A) or structurally enforcing commutative token aggregation (Intervention B)—the model will bypass the prolonged memorization phase entirely.

\paragraph{Modular Multiplication.}
To evaluate whether the acceleration generalizes beyond addition, we also train on modular multiplication $(a \times b) \bmod p$ with $p = 113$. We use an identical dataset format to addition: all $p^2$ input pairs (including zero) with a 30\% train / 70\% test split. For prime $p$, the non-trivial multiplicative structure is governed by the cyclic group $\mathbb{Z}_p^*$ of order $p-1 = 112$, whose representations admit the same discrete Fourier decomposition as the additive task. Following \citet{prieto2025grokking}, all multiplication configurations---including baselines---use a normalized unembedding matrix with fixed temperature scaling to prevent Softmax Collapse, isolating the effect of the residual stream topology.

\subsection{Task 2: The Symmetric Group $S_5$ (Negative Control)}
\label{sec:s5_task}

To isolate the effect of task-specific geometric alignment from generic optimization stabilization, we employ the composition of the symmetric group $S_5$ as a negative control. The $S_5$ composition task was originally introduced as a benchmark for delayed generalization by \citet{power2022grokking}. The dataset consists of the pairwise composition of the 120 permutations of five elements ($a \circ b = c$). Similar to the modular addition task, we format the input sequences as \texttt{[a, b, equals\_token]} and enforce a strict data scarcity regime by utilizing a random $30\%$ split for training ($4,320$ samples) and $70\%$ for testing ($10,080$ samples).

Unlike modular addition, $S_5$ is strictly non-commutative ($a \circ b \neq b \circ a$). Prior interpretability analyses of transformers trained on this task indicate that successful models construct inherently non-abelian representation structures—whether through higher-dimensional irreducible representations \cite{chughtai2023neural} or subgroup coset circuits \cite{stander2024grokking}—rather than relying on low-dimensional circular manifolds \cite{nanda2023progress}. In contrast to the cyclic group $\mathbb{Z}_p$, whose representations reduce to 1D Fourier modes lying on a circle, empirical analyses suggest that successful solutions to $S_5$ composition typically involve architectures capable of supporting higher-dimensional, non-commutative structure.

Importantly, we do not claim that a spherical constraint is fundamentally incompatible with permutation composition. Rather, our hypothesis is empirical: if the acceleration observed in modular addition arises from geometric alignment between the spherical constraint and the task’s intrinsic symmetry structure, then imposing the same constraint on $S_5$—whose learned solutions are typically higher-dimensional and non-commutative—should not yield similar acceleration.

Specifically, if the spherical topology functions merely as a generic regularizer or stabilizer, it should reduce the grokking delay across tasks. Conversely, if acceleration depends on task-specific alignment, then constraining representations to a fixed-norm spherical manifold may restrict the model's ability to construct the higher-dimensional structures typically observed in successful $S_5$ solutions, leading to delayed or absent generalization under the same training regime.

Demonstrating such a differential outcome would provide further support for our central hypothesis: that eliminating or shortening the memorization phase depends on aligning architectural degrees of freedom with the mathematical symmetries most naturally exploited by the task.

\section{Empirical Results}
\label{sec:results}

In this section, we evaluate the hypothesis that excess representational degrees of freedom—specifically, residual magnitude flexibility and data-dependent attention routing—contribute to delayed generalization in modular arithmetic tasks. We evaluate our two structural interventions across standard baselines (LayerNorm, RMSNorm) and our bounded spherical topology tested with weight decay ($\lambda=1.0$) and without ($\lambda=0.0$). All reported metrics are aggregated across 10 independent random seeds per configuration.

\paragraph{Experimental Setup.}
Across all configurations, we employ a shallow Transformer architecture with $d_{\text{model}} = 128$, $4$ attention heads, and an MLP hidden dimension of $512$. Models are trained using full-batch gradient descent with the AdamW optimizer. Depending on the intervention, we evaluate learning rates of $1 \times 10^{-4}$ and $6 \times 10^{-4}$ alongside weight decay values of $\lambda=1.0$ or $\lambda=0.0$. Comprehensive hyperparameter configurations, including the specific adjustments for the $S_5$ task, are detailed in Appendix \ref{app:hyperparameters}.

\subsection{Intervention A: Accelerating Grokking via Topological Constraint}
\label{sec:results_acceleration}

We first evaluate generalization dynamics under standard learned attention at a stable learning rate ($10^{-4}$). As shown in Table~\ref{tab:main_results}, both LayerNorm and RMSNorm baselines exhibit characteristic grokking behavior: models rapidly achieve perfect training accuracy but require an extended optimization plateau before test accuracy rises, with mean generalization epochs of 54,160 and 51,240 respectively.

\begin{figure}[htbp]
    \centering
    \begin{subfigure}[b]{0.48\textwidth}
        \centering
        \includegraphics[width=\textwidth]{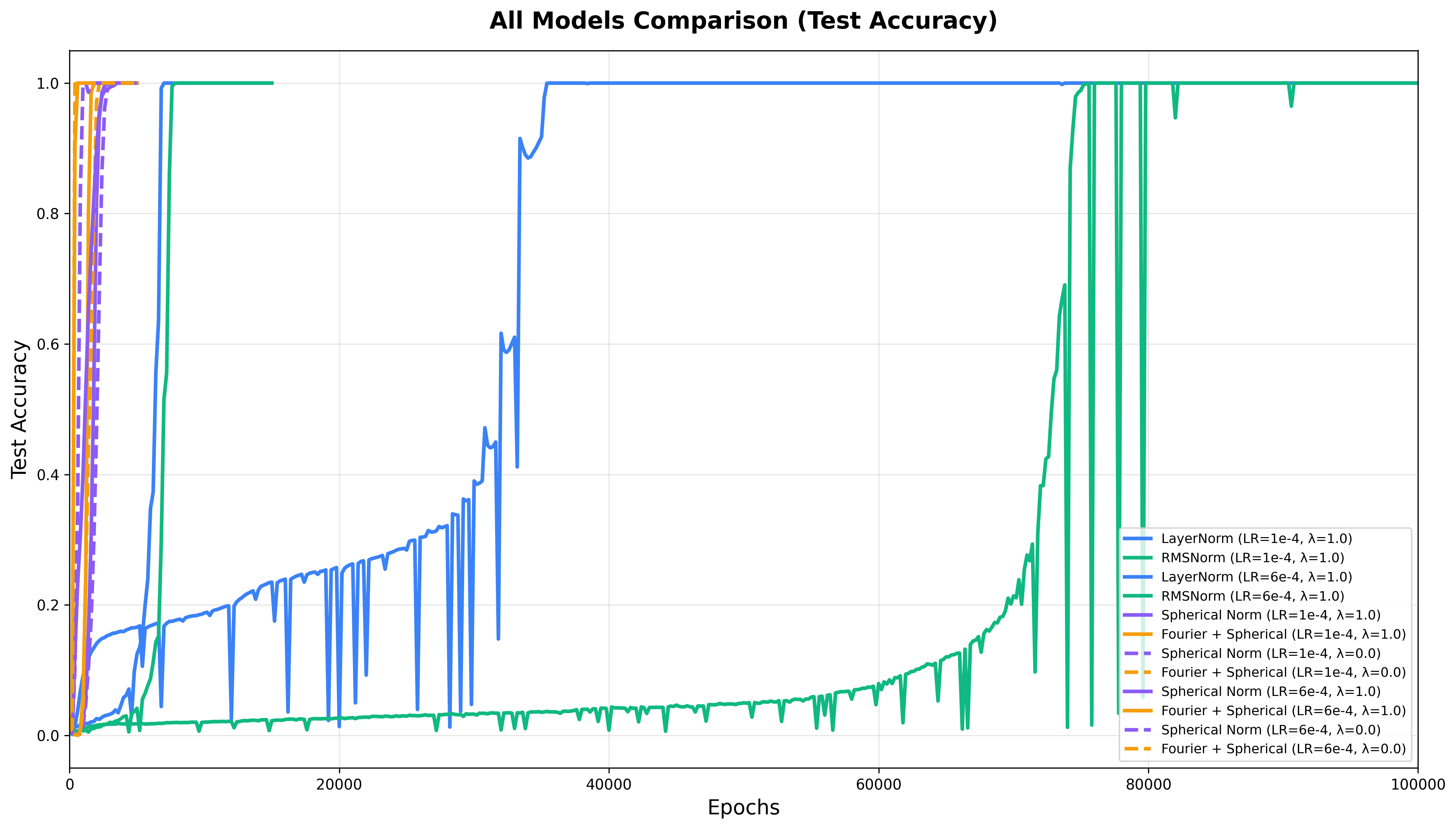}
        \caption{Macro Dynamics (100,000 Epochs)}
        \label{fig:macro_dynamics}
    \end{subfigure}
    \hfill
    \begin{subfigure}[b]{0.48\textwidth}
        \centering
        \includegraphics[width=\textwidth]{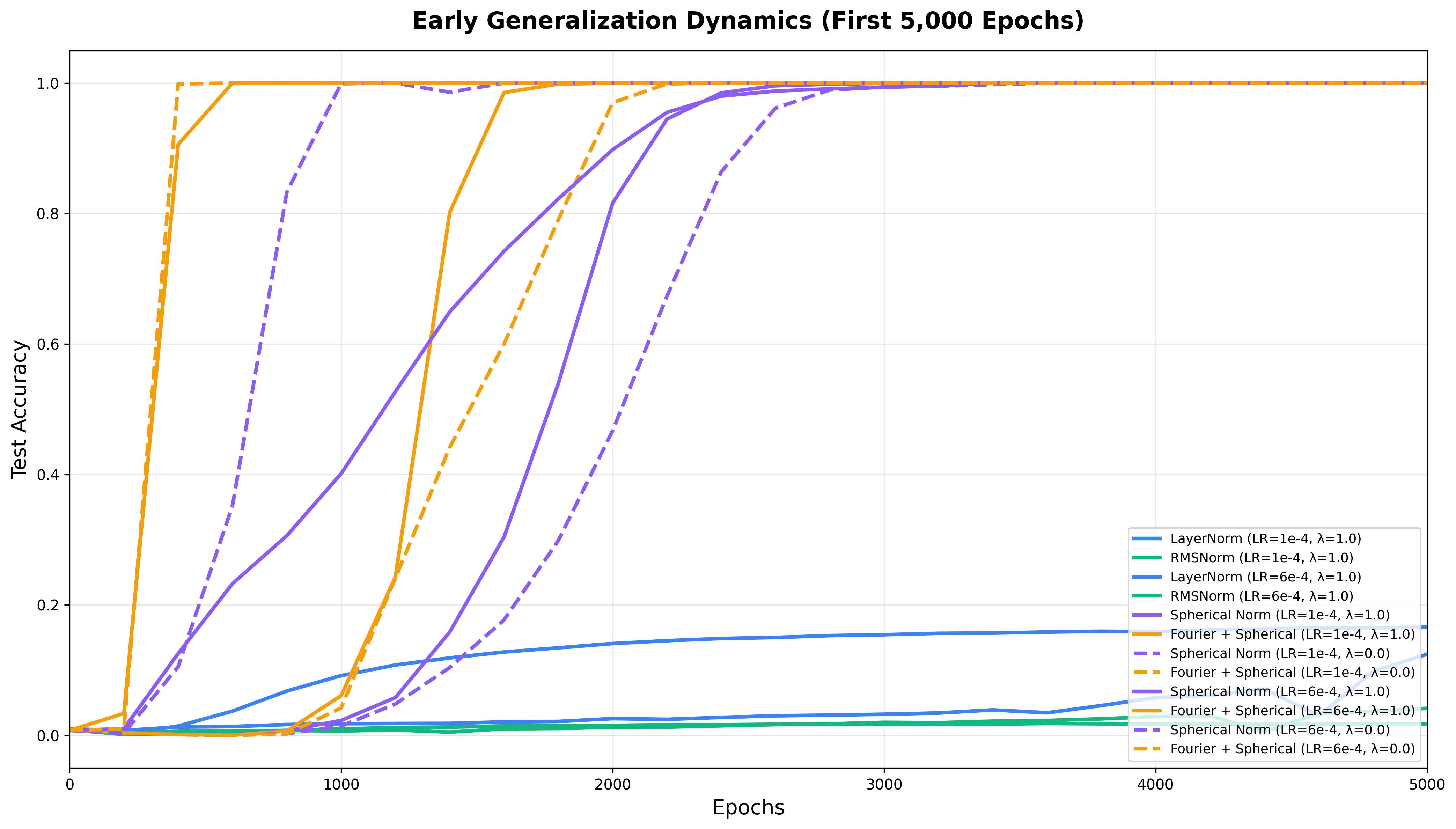}
        \caption{Early Dynamics (First 5,000 Epochs)}
        \label{fig:micro_dynamics}
    \end{subfigure}
    
    \caption{Test accuracy across architectural configurations. \textbf{(a)} Standard normalizations exhibit the classic, delayed grokking profile over 100,000 epochs. \textbf{(b)} A zoomed-in view of the first 5,000 epochs reveals that the topologically bounded models bypass this delay entirely, exhibiting immediate and stable convergence while the baselines remain trapped at chance accuracy.}
    \label{fig:all_models_comparison}
\end{figure}

Inspection of the overall learning trajectories (visualized in Figure \ref{fig:all_models_comparison}) reveals the stark contrast between unconstrained and topologically bounded models. During the extended optimization plateau, the baseline models exhibit pronounced oscillations in test accuracy—intermittent gradient spikes and transient drops in generalization before eventual stabilization—consistent with the ``slingshot effect'' reported in prior work \cite{thilak2022slingshot, prieto2025grokking}. To avoid masking this behavior, gradients were not clipped during training. By contrast, the spherical and fully bounded architectures (clustered at the extreme left of Figure \ref{fig:all_models_comparison}) entirely bypass these chaotic optimization dynamics, locking into the generalizing solution smoothly and immediately.

\begin{table}[htbp]
\centering
\caption{Grokking onset epoch across 10 random seeds. Spherical configurations substantially reduce convergence time at both standard and elevated learning rates.}
\label{tab:main_results}
\resizebox{\textwidth}{!}{
\begin{tabular}{llcccccc}
\toprule
\textbf{Learning Rate} & \textbf{Architecture} & \textbf{Mean Grok Epoch} & \textbf{Std Dev} & \textbf{Min} & \textbf{Max} & \textbf{Failures} & \textbf{Peak Acc.} \\
\midrule
\multirow{6}{*}{$1\times 10^{-4}$} 
& LayerNorm (Baseline) & 54,160 & 13,490 & 32,800 & 71,600 & 0 / 10 & 100\% \\
& RMSNorm (Baseline) & 51,240 & 11,200 & 38,800 & 74,600 & 0 / 10 & 100\% \\
& Bounded Sphere ($\lambda=1.0$) & 2,400 & 353 & 1,800 & 3,000 & 0 / 10 & 100\% \\
& Bounded Sphere ($\lambda=0.0$) & 2,480 & 464 & 1,800 & 3,200 & 0 / 10 & 100\% \\
& Bounded Sphere + Fourier Init ($\lambda=1.0$) & 2,120 & 368 & 1,600 & 2,600 & 0 / 10 & 100\% \\
& \textbf{Bounded Sphere + Fourier Init ($\lambda=0.0$)} & \textbf{2,100} & \textbf{316} & \textbf{1,800} & \textbf{2,600} & \textbf{0 / 10} & \textbf{100\%} \\
\midrule
\multirow{6}{*}{$6\times 10^{-4}$} 
& LayerNorm (Baseline) & 7,800 & 1,095 & 6,000 & 9,400 & 0 / 10 & 100\% \\
& RMSNorm (Baseline) & 7,300 & 925 & 6,000 & 9,200 & 0 / 10 & 100\% \\
& Bounded Sphere ($\lambda=1.0$) & 1,560 & 853 & 600 & 3,000 & 0 / 10 & 100\% \\
& Bounded Sphere ($\lambda=0.0$) & 820 & 199 & 600 & 1,200 & 0 / 10 & 100\% \\
& Bounded Sphere + Fourier Init ($\lambda=1.0$) & 800 & 163 & 600 & 1,000 & 0 / 10 & 100\% \\
& \textbf{Bounded Sphere + Fourier Init ($\lambda=0.0$)} & \textbf{711} & \textbf{203} & \textbf{400} & \textbf{1,000} & \textbf{0 / 10} & \textbf{100\%} \\
\bottomrule
\end{tabular}%
}
\end{table}

In contrast to the baselines, the spherical configurations exhibit substantially earlier generalization. At $10^{-4}$, the Bounded Sphere ($\lambda=0.0$) reaches 100\% test accuracy in a mean of 2,480 epochs—over an order of magnitude faster than the statistical normalization baselines. At a higher learning rate ($6\times 10^{-4}$), baselines require approximately 7,500 epochs on average, while the Bounded Sphere ($\lambda=0.0$) generalizes in roughly 820 epochs. 

\paragraph{Empirical Validation of the Bounded Topology.}
As theorized in Section~\ref{sec:spherical}, standard unconstrained models rely on explicit regularization to counteract Naïve Loss Minimization and prevent Softmax Collapse \cite{prieto2025grokking}. All spherical configurations employ the same bounded architecture: $L_2$-normalized residual stream, normalized unembedding, and fixed temperature scaling. The only variable is weight decay. At the lower learning rate ($1 \times 10^{-4}$), weight decay has minimal effect—both $\lambda=1.0$ and $\lambda=0.0$ converge in $\sim$2,400--2,480 epochs. At the higher learning rate ($6 \times 10^{-4}$), however, weight decay introduces optimization friction: the $\lambda=1.0$ configuration exhibits higher variance across seeds (std 853 vs.\ 199) and slower mean convergence (1,560 vs.\ 820 epochs). This suggests that under the bounded topology, weight decay penalizes weight magnitudes that are already geometrically constrained, introducing unnecessary regularization pressure at aggressive learning rates.

Crucially, bounding the unembedding alone does not eliminate delayed generalization. When an $L_2$-normalized unembedding is applied to the LayerNorm baseline, the model still exhibits a prolonged memorization phase. This occurs because LayerNorm preserves internal radial freedom in the residual stream, allowing the network to increase residual magnitudes in order to reduce cross-entropy loss through Na\"{i}ve Loss Minimization \cite{prieto2025grokking}. While recent work proposes numerical stabilizations (e.g., StableMax) that allow models to remain numerically stable during this unconstrained scaling phase \cite{prieto2025grokking}, our bounded spherical topology instead removes this internal magnitude degree of freedom at the architectural level. By structurally constraining the residual stream itself, the architecture bypasses the prolonged memorization phase observed in baseline models and exhibits immediate generalization.

\paragraph{Extension to Modular Multiplication.}
To confirm that the acceleration is not specific to addition, we evaluated all configurations on modular multiplication $(a \times b) \bmod 113$. As shown in Table~\ref{tab:mult}, the bounded spherical topology produces comparable acceleration. At the 95\% generalization milestone, Bounded Sphere models reach generalization in $\sim$2,200 epochs compared to $\sim$28,000--31,000 for baselines—a $\sim$13$\times$ speedup. All bounded sphere seeds ultimately reach 100\% test accuracy.

Notably, the spectral verification confirms that multiplication models learn the same Fourier-based circuit: causal ablation retaining only the top 5 frequencies preserves $>$99.75\% accuracy across all bounded sphere seeds.

 
\begin{table}[h]
\centering
\caption{Generalization dynamics for modular multiplication $(a \times b) \bmod 113$ across 10 random seeds. We report the epoch at which each seed first reaches 95\% test accuracy as the primary generalization milestone. The bounded spherical topology accelerates generalization comparably to the addition task.}
\label{tab:mult}
\small
\begin{tabular}{lccccc}
\toprule
Architecture & Mean 95\% Epoch & Std Dev & Min & Max & Hits \\
\midrule
LayerNorm (Baseline) & 28,220 & 3,676 & 19,600 & 32,400 & 10\,/\,10 \\
RMSNorm (Baseline) & 31,360 & 4,056 & 25,600 & 38,600 & 10\,/\,10 \\
Bounded Sphere ($\lambda = 1.0$) & 2,320 & 223 & 2,000 & 2,800 & 10\,/\,10 \\
Bounded Sphere ($\lambda = 0.0$) & 2,200 & 253 & 1,800 & 2,600 & 10\,/\,10 \\
\bottomrule
\end{tabular}
\end{table}

\subsection{Intervention B: Bypassing Grokking via Uniform Attention Ablation}
\label{sec:results_attention}

Independent of magnitude constraints, we tested whether the capacity for data-dependent query-key routing contributes to the observed grokking delay. Based on theoretical proofs that modular addition requires only a uniform ``bag-of-tokens'' aggregation \cite{huang2025provable}, we applied the Uniform Attention Ablation (forcing attention weights to a static $[1/3, 1/3, 1/3]$). Models were trained for 20,000 epochs to definitively evaluate whether they could escape the memorization basin without adaptive routing. 

\begin{table}[htbp]
\centering
\caption{Peak test accuracy under Uniform Attention Ablation (Zero-Attention) across 10 random seeds. Ablating data-dependent routing allows normalized models to completely bypass the grokking delay.}
\label{tab:zero_attention_results}
\resizebox{\textwidth}{!}{
\begin{tabular}{lccc}
\toprule
\textbf{Architecture (Uniform Attention)} & \textbf{Mean Peak Acc.} & \textbf{Max Peak Acc.} & \textbf{Success Rate (100\% Acc)} \\
\midrule
\textbf{LayerNorm Baseline (WD=1.0)} & \textbf{100.00\%} & \textbf{100.00\%} & \textbf{10 / 10} \\
Bounded Sphere ($\lambda=1.0$) & 97.65\% & 100.00\% & 6 / 10 \\
\textbf{Bounded Sphere ($\lambda=0.0$)} & \textbf{100.00\%} & \textbf{100.00\%} & \textbf{10 / 10} \\
\bottomrule
\end{tabular}%
}
\end{table}

As shown in Table~\ref{tab:zero_attention_results}, the standard LayerNorm baseline equipped with uniform attention achieves 100\% peak test accuracy across all 10 independent seeds, entirely bypassing the prolonged grokking plateau. The Bounded Sphere ($\lambda=0.0$) likewise maintains 100\% success under uniform attention. 

Beyond peak accuracy, examining the training dynamics under uniform attention reveals the impact of topological constraints on optimization stability. As illustrated in Figure~2, both the standard LayerNorm baseline (Figure~2a) and the Bounded Sphere ($\lambda=0.0$) (Figure~2c) exhibit an immediate, concurrent rise in training and test accuracy, bypassing the extended memorization plateau observed in other settings. In contrast, when the spherical constraint is applied while retaining weight decay (Figure~2b), the optimization trajectory becomes markedly unstable. The tension between the model increasing logit magnitudes to reduce cross-entropy loss and weight decay penalizing that growth appears to introduce substantial optimization friction, resulting in delayed generalization.

These observations suggest that removing magnitude-based representational freedom—via a fully bounded manifold combined with zero weight decay—can promote stable, phase-free convergence in this task setting. 

More broadly, the results indicate that complex, data-dependent routing is not strictly necessary for this commutative operation. By reducing the attention mechanism to a Continuous Bag-of-Words-style uniform aggregator, we eliminate symmetry-breaking routing pathways, which in this setting leads to immediate generalization.

\begin{figure}[htbp]
    \centering
    \begin{subfigure}[b]{0.32\textwidth}
        \centering
        \includegraphics[width=\textwidth]{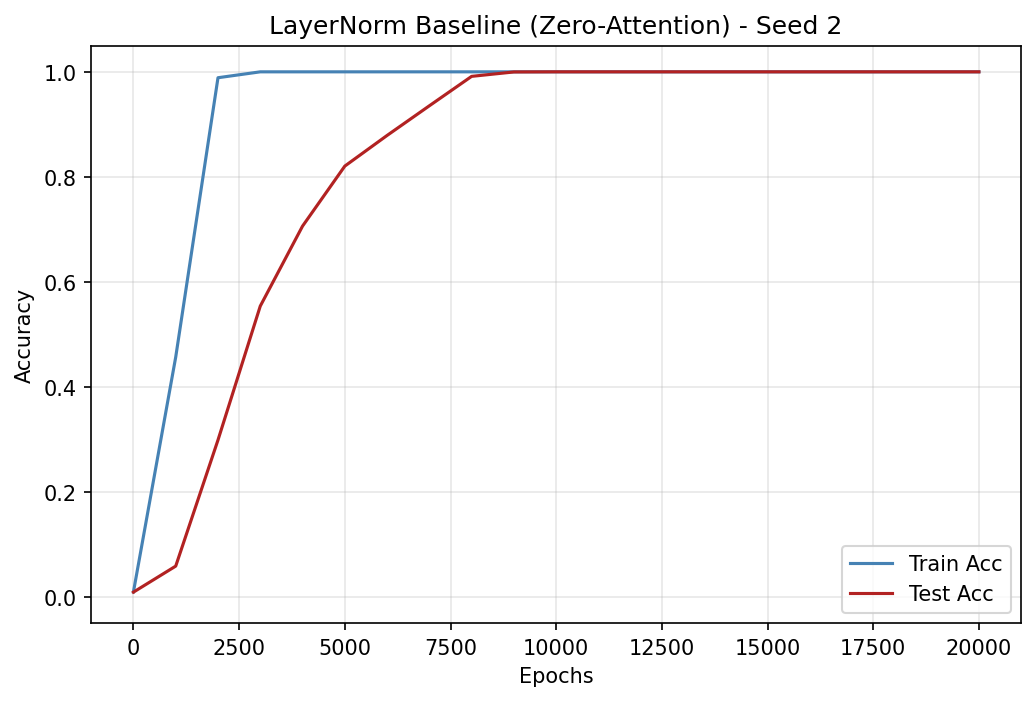}
        \caption{LayerNorm (Baseline)}
        \label{fig:zero_attn_layernorm}
    \end{subfigure}
    \hfill
    \begin{subfigure}[b]{0.32\textwidth}
        \centering
        \includegraphics[width=\textwidth]{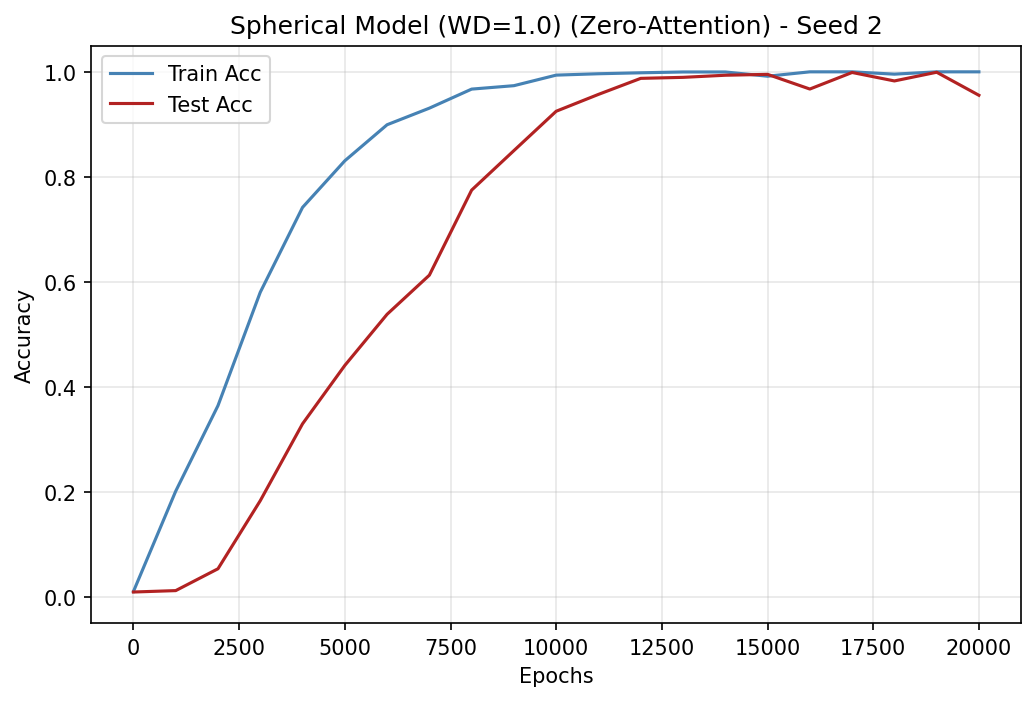}
        \caption{Bounded Sphere ($\lambda=1.0$)}
        \label{fig:zero_attn_sphere_wd}
    \end{subfigure}
    \hfill
    \begin{subfigure}[b]{0.32\textwidth}
        \centering
        \includegraphics[width=\textwidth]{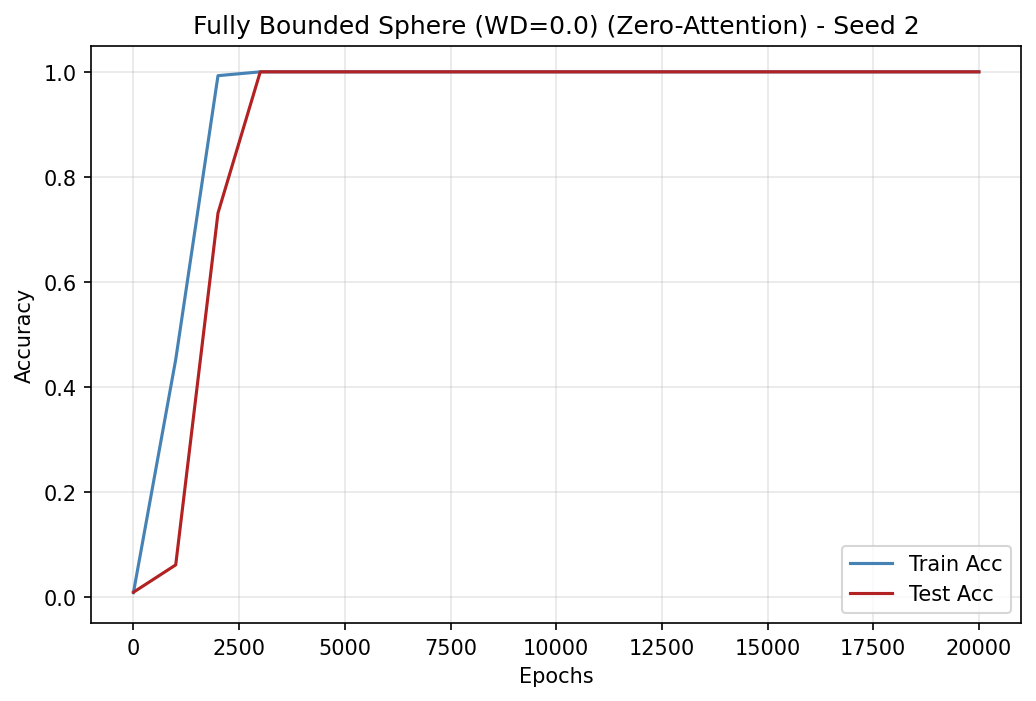}
        \caption{Bounded Sphere ($\lambda=0.0$)}
        \label{fig:zero_attn_fully_bounded}
    \end{subfigure}
    
    \caption{Training dynamics under Uniform Attention Ablation (Seed 2). \textbf{(a)} The LayerNorm baseline immediately generalizes when data-dependent routing is removed, bypassing grokking. \textbf{(b)} Imposing a spherical constraint while retaining weight decay introduces optimization instability, delaying generalization. \textbf{(c)} The Bounded Sphere without weight decay ($\lambda=0.0$) stabilizes the geometry, resulting in immediate and flawless generalization.}
    \label{fig:zero_attention_dynamics}
\end{figure}

\subsection{Spectral Verification of the Fourier Circuit}\label{sec:spectral}
\label{sec:results_spectral}

While test accuracy confirms successful generalization, it does not reveal \textit{how} the models solve the task. Prior mechanistic interpretability work shows that transformers trained on modular addition construct a Fourier circuit, mapping discrete inputs onto a continuous 1D circular manifold \cite{nanda2023progress}.

To evaluate whether our topological constraints preserve or distort this structure, we performed a spectral analysis of the learned representations. Our goal is to determine whether acceleration arises from earlier construction of the canonical Fourier circuit or from an alternative representational strategy. We constructed the effective linear transformation $W_L = W_U W_{out}$ and identified the top five highest-magnitude frequencies via Fast Fourier Transform (FFT). We then conducted an ablation by restricting the logits to these top five frequencies. Across successful configurations, this ablation preserved $>99\%$ test accuracy (e.g., 99.98\% for LayerNorm, 99.96\% for Fully Bounded), indicating that these components strictly dominate the generalization mechanism. This confirms that the accelerated models rely on the same Fourier-based solution identified in prior work, rather than an alternative shortcut mechanism.

To quantify geometric structure, we computed the Fraction of Variance Explained (FVE). For each dominant frequency $k$, we measured the proportion of variance in the MLP activations explained by the ideal 2D basis functions $\cos(\omega_k(a+b))$ and $\sin(\omega_k(a+b))$. 

\paragraph{Baseline Circuit.}
The LayerNorm baseline constructs a recognizable Fourier circuit, with dominant frequencies explaining up to $\sim 54\%$ of the variance in MLP activations. In successful runs, coherent Fourier structure is observed after the prolonged optimization plateau described in Section~\ref{sec:results_acceleration}.

\paragraph{Optimization Friction (Bounded Sphere, $\lambda=1.0$).}
When the bounded spherical topology is trained with standard weight decay ($\lambda=1.0$), circuit coherence degrades substantially. The maximum FVE does not exceed $\sim 29\%$, with several dominant frequencies explaining less than 10\% of activation variance. This fragmentation is consistent with the optimization instability arising from the interaction between loss-driven logit scaling and weight decay regularization.
This instability admits a precise interpretation through the concurrent
Norm-Separation Delay Law of Truong et al.~\cite{truong2026norm}, whose
dynamical analysis shows that weight decay contracts parameter norms by a
factor of $(1 - \eta\lambda)$ per step.  Under the bounded spherical topology,
this contraction force conflicts directly with the $\Pi_S$ projection
operator, which rigidly rescales activations back to unit norm---producing
the oscillatory dynamics and degraded circuit coherence observed above.
We expand on this connection in Section~\ref{sec:discussion}.

\paragraph{Bounded Sphere ($\lambda=0.0$).}
In contrast, the same bounded topology trained without weight decay ($\lambda=0.0$) exhibits strong spectral alignment. Dominant frequencies explain a substantial fraction of activation variance (e.g., FVE U: 62.55\%, V: 48.43\%). By removing magnitude scaling from both the residual stream and the unembedding layer, optimization converges rapidly to a geometrically coherent Fourier representation.

\subsection{Task-Specific Alignment: Negative Control on $S_5$}
\label{sec:results_s5}

To determine whether the acceleration observed in Intervention A (the Fully Bounded topology) is the result of a generic optimization stabilizer or a task-specific geometric inductive bias, we evaluated the architectures on the composition of the symmetric group $S_5$. As detailed in Section~\ref{sec:s5_task}, $S_5$ is non-commutative and requires higher-dimensional representational spaces that do not reduce to the continuous 1D circular manifold governing modular addition. 

Because this task is fundamentally more complex, all models failed to converge within 100,000 epochs at previous learning rates; therefore, we trained all models for up to 100,000 epochs at an elevated learning rate of $10^{-3}$. The results are summarized in Table~\ref{tab:s5_results}.

\begin{table}[htbp]
\centering
\caption{Grokking onset epoch for the $S_5$ permutation composition task. Metrics for baselines are calculated exclusively over successful seeds. While the baselines generalize on the majority of seeds, the strict spherical constraints fail to produce generalization within the 100,000-epoch training window.}
\label{tab:s5_results}
\resizebox{0.95\textwidth}{!}{
\begin{tabular}{lccccc}
\toprule
\textbf{Architecture} & \textbf{Mean Grok Epoch} & \textbf{Std Dev} & \textbf{Min} & \textbf{Max} & \textbf{Failures} \\
\midrule
LayerNorm (Baseline) & 39,900 & 11,942 & 24,400 & 58,400 & 2 / 10 \\
RMSNorm (Baseline) & 38,250 & 23,600 & 19,000 & 91,600 & 2 / 10 \\
Bounded Sphere ($\lambda=1.0$) & \textbf{Failed} & \textbf{--} & \textbf{--} & \textbf{--} & \textbf{10 / 10} \\
\textbf{Bounded Sphere ($\lambda=0.0$)} & \textbf{Failed} & \textbf{--} & \textbf{--} & \textbf{--} & \textbf{10 / 10} \\
\bottomrule
\end{tabular}%
}
\end{table}

Crucially, while the standard baselines successfully grok $S_5$ (mean $\sim$40,000 epochs on successful runs), the bounded spherical topologies fail to achieve generalization on any seed within the 100,000-epoch training window. Despite reaching 100\% training accuracy, the spherical models remain confined to a memorization plateau, with test accuracy remaining near chance throughout training.

This differential outcome provides evidence that the bounded spherical topology functions as a task-specific geometric inductive bias rather than a generic optimization accelerator. If the spherical constraint merely improved optimization dynamics, comparable acceleration would be expected across tasks. Instead, we observe acceleration on modular addition but consistent failure on $S_5$ under the same training regime.

These results support the hypothesis that shortening delayed generalization depends on alignment between architectural degrees of freedom and the symmetry structure most naturally exploited by the task. In the commutative case ($\mathbb{Z}_p$), the spherical constraint aligns with the circular Fourier geometry underlying successful solutions. In contrast, for the non-commutative $S_5$ task—where prior analyses indicate higher-dimensional representation structure—the same constraint appears to hinder the construction of generalizing circuits in this experimental setting.

We additionally evaluated the Uniform Attention Ablation on $S_5$. Standard baselines with uniform attention achieve $\sim$99\% peak accuracy but do not exhibit meaningful acceleration relative to learned attention, confirming that data-dependent routing is not the primary bottleneck for non-commutative tasks. Spherical models with uniform attention fail entirely ($\sim$2.4\% accuracy), consistent with the learned-attention results. Full results are reported in Appendix~\ref{app:s5_cbow}.

\section{Discussion and Conclusion}\label{sec:discussion}
\label{sec:conclusion}

\subsection{From Post-Hoc Analysis to Predictive Intervention}

Mechanistic interpretability has traditionally functioned as a post-hoc observational science: models are trained under standard architectural assumptions, and circuits are subsequently reverse-engineered to explain emergent behavior. In contrast, this work demonstrates an interventional methodology. Building on prior findings that grokking in modular addition coincides with the emergence of Fourier representations, we directly restricted the architecture's degrees of freedom to better align with the known Fourier structure of the task prior to training.

By imposing an $L_2$ spherical constraint (Intervention A), we structurally aligned the residual stream with the continuous circular manifold required for Fourier features, substantially accelerating generalization (over an order of magnitude in our setting). Similarly, by ablating data-dependent routing and enforcing uniform token aggregation (Intervention B), we matched the theoretical commutative requirements of the task, eliminating the delayed memorization phase in this setting. This bidirectional result—further validated by the failure of the spherical constraint on the non-commutative $S_5$ task—provides strong interventional evidence that architectural topology can function not merely as a descriptive lens, but as a predictive probe of task–model alignment. Representation analysis informs architectural intervention, and architectural intervention tests mechanistic hypotheses.

\paragraph{Connection to the Norm-Separation Delay Law.}

In concurrent independent work, Truong et al.~\cite{truong2026norm} derive tight
bounds on the grokking delay, establishing that
$T_{\mathrm{grok}} - T_{\mathrm{mem}}
  = \Theta\!\bigl(\gamma_{\mathrm{eff}}^{-1}\,
    \log \tfrac{\|\theta_{\mathrm{mem}}\|^{2}}
               {\|\theta_{\mathrm{post}}\|^{2}}\bigr)$,
where $\gamma_{\mathrm{eff}}$ is the effective contraction rate under weight
decay.  Their framework decomposes the grokking delay into two phases:
an \emph{optimization escape} phase, during which weight decay exponentially
contracts parameter norms from a high-norm memorization state toward a 
low-norm Fourier manifold, and a subsequent \emph{statistical
confirmation} phase in which validation accuracy rises sharply.
This decomposition provides a precise quantitative lens through which
to interpret our architectural interventions.

\paragraph{Structural elimination of the norm-separation precondition.}

Theorem~3.8 of Truong et al.\ establishes that norm separation---specifically,
$\|\theta_{\mathrm{mem}}\|^{2} > \|\theta_{\mathrm{post}}\|^{2}$---is a
\emph{necessary} condition for any positive grokking delay under regularized
first-order dynamics.  Our Fully Bounded Spherical Topology acts directly on
this precondition.  By applying the projection operator $\Pi_S$ throughout the
residual stream and constraining the unembedding matrix to unit norm with a
fixed temperature scale, the architecture removes the representational capacity
to construct high-norm memorization interpolants.  Although $\Pi_S$ constrains
only the residual stream activations (not the full parameter vector
$\theta$), the elimination of magnitude as a representational axis
prevents the network from encoding training-specific information in
vector norm---precisely the mechanism that, in unconstrained
architectures, produces high-norm memorization solutions.  The empirical 
consequence is that the norm gap $\log(\|\theta_{\mathrm{mem}}\|^2 / \|\theta_{\mathrm{post}}\|^2)$ 
collapses, structurally eliminating the optimization escape phase that 
constitutes the bulk of the grokking delay.  This is consistent with Truong et 
al.'s necessity theorem: when the architectural capacity for norm separation 
is absent, no delayed transition is possible---which is exactly what we 
observe when bounded models generalize without a memorization phase, 
even at $\lambda = 0$.

\paragraph{Explaining the optimization friction under weight decay.}

The interaction between our spherical constraint and weight decay
($\lambda = 1.0$) also admits a clean explanation through Truong et al.'s
dynamical framework.  Their analysis shows that regularized SGD contracts
parameter norms by a factor of $(1 - \eta\lambda)$ per step
(Equation~1 of~\cite{truong2026norm}).  Under the bounded spherical topology,
however, $\Pi_S$ rigidly rescales activations back to unit norm after each
sub-layer.  When weight decay is active, a mathematical conflict arises: regularized SGD
continuously shrinks the network's parameters. However, because $\Pi_S$ rigidly
rescales the forward-pass activations to unit norm, the model's output becomes
scale-invariant. As weight decay drives the parameter norms to artificially
small values without increasing the loss, the effective learning rate---which
scales inversely with parameter magnitude in scale-invariant networks---explodes.
This angular instability explains the optimization friction we observe in
Section~\ref{sec:spectral} (Bounded Sphere, $\lambda = 1.0$), where circuit
coherence degrades substantially (maximum FVE $\sim$29\%) and
cross-seed variance increases.  In contrast, removing weight decay
($\lambda = 0.0$) eliminates this conflict entirely: the bounded topology
is free to stabilize on the Fourier manifold without the competing
contraction signal, producing the strong spectral alignment and immediate
generalization we report (FVE U: 62.55\%, V: 48.43\%).

\paragraph{Convergence from opposite directions.}

Together, the two works converge on a single picture from complementary
directions.  Truong et al.\ prove that the magnitude gap between
memorization and generalization solutions is what \emph{determines} the time
scale of grokking under regularized optimization; we demonstrate empirically
that \emph{removing} the architectural capacity for that magnitude gap
collapses the delay.  The theory predicts the intervention's effect; the
intervention validates the theory's mechanism.  Our S5 negative control
further reinforces this convergence: the norm-separation structure exploited
by the Delay Law is task-specific, depending on memorizing and generalizing
solutions occupying distinct norm regimes.  For tasks whose generalizing
representations are not low-norm relative to memorizing ones---as with the
discrete coset structures required for S5 permutation
composition~\cite{stander2024grokking}---architecturally enforcing a
spherical geometry should not be expected to reduce the delay, which is
exactly what we observe.

\paragraph{From interpretation to design.}
Our results demonstrate a closed loop between mechanistic interpretability and architectural design. Prior work identified Fourier circuits as the generalizing solution for modular arithmetic \citep{nanda2023progress}; we used this knowledge to design an architectural constraint aligned with that solution; the constraint eliminated the memorization phase. This pipeline---\emph{interpret circuits, then design architecture}---converts post-hoc understanding into engineering leverage, even at toy scale. Importantly, the pipeline also operates in reverse: when the underlying circuit is \emph{not} known, testing architectural constraints and observing which accelerate or impede generalization can reveal properties of the task's intrinsic structure. Our $S_5$ result illustrates this---the failure of the spherical constraint is itself informative, indicating that the generalizing representation does not reside on a continuous circular manifold, consistent with the discrete coset structures identified by \citet{stander2024grokking}. The natural next candidates for this bidirectional approach are domains with plausible but unconfirmed periodic or symmetric structure where mechanistic interpretability remains underexplored, such as time-series forecasting, signal processing, or physical simulation. For heterogeneous domains like natural language, however, the pipeline faces a fundamental obstacle: the diversity of latent structures makes it unlikely that any single architectural prior can serve all subtasks simultaneously.
We additionally evaluated $L_1$ normalization, which constrains representations to a cross-polytope geometry. This modification also failed to produce generalization on $S_5$, suggesting that the task's coset-based structure is fundamentally incompatible with simple norm-based topological constraints.
\subsection{Toward Task-Specific Structural Alignment}

Our results do not imply the existence of a single universal architectural bias. Rather, they suggest a task-by-task research program: (1) identify the representation and routing structure that naturally emerges during successful generalization, (2) encode this structure as an architectural prior, and (3) evaluate whether the memorization-to-generalization transition is shortened or eliminated.

The negative result on $S_5$ serves a critical complementary role. Because the spherical constraint is not aligned with the non-abelian representation structure underlying permutation composition, it fails to produce generalization under the same training setup. This contrast reinforces the hypothesis that delayed generalization is highly sensitive to the alignment between architectural degrees of freedom and task symmetry, rather than being solely an unavoidable optimization artifact. Extending alignment to non-abelian or hierarchical symmetry structures remains an open and technically nontrivial direction for future work.

\subsection{The Bitter Lesson and Modality-Specific Debugging}

For domains such as natural language, the underlying symmetry structure may be heterogeneous, hierarchical, or only approximately harmonic. In such cases, hard-coding a single global geometric prior is unlikely to suffice. This perspective does not contradict Sutton's \textit{Bitter Lesson} \citep{sutton2019bitter}; rather, it reframes structural alignment as a diagnostic and interventional tool for domains where mathematical structure is known or controllable.

Transformers are increasingly deployed beyond text, including in time-series forecasting, reinforcement learning, and structured decision-making tasks. In such settings, recent architectural interventions provide suggestive empirical support for symmetry-aligned design. For example, \citet{gillman2025fourier} demonstrate that replacing a standard linear classification head with a topologically constrained ``Fourier head'' significantly improves performance on continuous decision-making tasks. Similarly, \citet{sun2025penguin} show that explicitly incorporating a modulo-based periodic relative attention bias in the PENGUIN architecture yields strong results in long-horizon time-series forecasting by structurally aligning the attention mechanism with the data's intrinsic cyclic symmetry.

While these works do not explicitly analyze grokking dynamics, they are consistent with the hypothesis advanced here: when architectural components are aligned with the intrinsic mathematical structure of a task, models may be able to bypass memorization-heavy regimes and more directly construct invariant representations.

Recent work has also begun extending grokking beyond synthetic algorithmic tasks into real-world reasoning domains. \citet{abramov2025grokking} demonstrate that augmenting sparse knowledge graphs with synthetic relational data can induce grokking-like phase transitions in multi-hop factual reasoning benchmarks. Viewed through our structural lens, such results reinforce the broader perspective that grokking reflects a transition from memorization to structured relational encoding. Our contribution complements this direction by showing that this transition need not be induced post hoc through data manipulation; when architectural degrees of freedom are restricted to match task symmetry, the delayed phase can disappear entirely.

Indeed, the success of nGPT \cite{loshchilov2025ngptnormalized} on language modeling may appear to contradict this conclusion. However, nGPT supplements its spherical constraint with learnable per-dimension scaling factors and eigen learning rates at every sublayer, effectively allowing the network to locally modulate or bypass the geometric restriction. This is architecturally distinct from the strict, uncompensated constraint we evaluate: our spherical topology offers no such escape mechanism, which is precisely why it serves as a diagnostic tool. The contrast between nGPT's flexible normalization succeeding on language and our rigid constraint failing on S5 reinforces rather than undermines our thesis: hard geometric priors accelerate learning only when aligned with task symmetry, while practical deployment on heterogeneous domains requires sufficient compensatory flexibility to accommodate misaligned subtasks.

\subsection{The Role of Synthetic Tasks and Scaling Geometric Bias}

While our empirical validation is conducted on synthetic algorithmic datasets, this is a deliberate methodological choice consistent with the foundational literature. The phenomenon of grokking itself was initially discovered and characterized within controlled environments. Similarly, the subsequent post-hoc recoveries of Fourier circuits, non-commutative coset structures (such as those required for $S_{5}$ permutation composition), and the recent identification of magnitude-driven Softmax Collapse were all achieved using mathematical toy tasks. These controlled settings are requisite for isolating structural phase transitions from the statistical noise of heterogeneous data. Consequently, our interventional methodology builds upon these environments to provide rigorous, mathematical validation of structural alignment. 

An important open question is whether these targeted constraints can improve performance beyond controlled symmetry tasks. A natural next step is to evaluate hybrid bounded-unconstrained architectures at moderate scale on heterogeneous corpora. If structural constraints primarily benefit domains with strong mathematical or cyclic symmetry, one would expect performance gains to concentrate in reasoning or algorithmic benchmarks while leaving purely linguistic benchmarks unaffected. However, such hybrid designs risk representation entanglement: without explicit routing regularizations, the model might incorrectly force unstructured linguistic features through the constrained spherical pathway. This could degrade overall performance and obfuscate post-hoc mechanistic analysis. 

\subsection{Conclusion}

We provide controlled interventional evidence that, in modular arithmetic tasks, grokking reflects a representational realignment process rather than solely an unavoidable optimization phase transition. When a model possesses excess architectural degrees of freedom—such as unbounded magnitude scaling or complex data-dependent routing—it can rely on memorization-heavy strategies before constructing structured representations. By isolating these degrees of freedom and restricting them to better match the intrinsic commutative and periodic symmetries of modular addition, we show that the prolonged generalization delay can be dramatically shortened or, in this experimental setting, eliminated.

More broadly, this work proposes a shift from post-hoc interpretability toward predictive structural debugging: a framework in which mechanistic representation analysis informs architectural design, and architectural design in turn provides experimental tests of interpretability hypotheses. While general-purpose AI systems will continue to rely on large, flexible architectures, task-specific structural alignment offers a principled pathway for isolating and studying the factors that influence generalization dynamics.

\section*{Acknowledgements}

Large language models were used in the preparation of this manuscript for limited assistance with language polishing, LaTeX formatting of tables and figures, and routine programming tasks. All experimental design, implementation, analysis, and scientific conclusions were developed and verified by the author, who takes full responsibility for the content of this work.

\section*{Code Availability}

All code necessary to reproduce the experiments in this work, including
model implementations, training scripts, hyperparameter configurations,
and analysis utilities for spectral verification and Fourier decomposition,
is publicly available at:
\url{https://github.com/AlperYildirim1/geometric-grokking}

\bibliographystyle{plainnat}
\bibliography{references} 

@inproceedings{nanda2023progress,
    title={Progress Measures for Grokking via Mechanistic Interpretability},
    author={Nanda, Neel and Chan, Lawrence and Lieberum, Tom and Smith, Jess and Steinhardt, Jacob},
    booktitle={The Eleventh International Conference on Learning Representations (ICLR)},
    year={2023},
    url={https://arxiv.org/abs/2301.05217}
}

@article{singh2026explaining,
    title={Explaining Grokking in Transformers through the Lens of Inductive Bias},
    author={Singh, Jaisidh and Misra, Diganta and Orvieto, Antonio},
    journal={arXiv preprint arXiv:2602.06702},
    year={2026}
}

@misc{stander2024grokking,
      title={Grokking Group Multiplication with Cosets}, 
      author={Dashiell Stander and Qinan Yu and Honglu Fan and Stella Biderman},
      year={2024},
      eprint={2312.06581},
      archivePrefix={arXiv},
      primaryClass={cs.LG}
}

@inproceedings{
zheng2024delays,
title={Delays in generalization match delayed changes in representational geometry},
author={Xingyu Zheng and Kyle Daruwalla and Ari S Benjamin and David Klindt},
booktitle={UniReps: 2nd Edition of the Workshop on Unifying Representations in Neural Models},
year={2024},
url={https://openreview.net/forum?id=1ae108kHk2}
}

@misc{minegishi2026emergent,
      title={Emergent Analogical Reasoning in Transformers}, 
      author={Gouki Minegishi and Jingyuan Feng and Hiroki Furuta and Takeshi Kojima and Yusuke Iwasawa and Yutaka Matsuo},
      year={2026},
      eprint={2602.01992},
      archivePrefix={arXiv},
      primaryClass={cs.AI},
      url={https://arxiv.org/abs/2602.01992}
}

@misc{lei2025construct,
      title={Construct-then-Compress: Geometric Dynamics of Grokking in Transformers}, 
      author={Ruian Lei and Zenglin Xu}, 
      year={2025},
      note={Published on OpenReview (Originally submitted as "Geometric Compression in Grokking")},
      url={https://openreview.net/} 
}

@inproceedings{huang2025provable,
  title={Provable Benefits of Sinusoidal Activation for Modular Addition},
  author={Tianlong Huang and Zhiyuan Li},
  booktitle={The Thirteenth International Conference on Learning Representations},
  year={2025},
  url={https://arxiv.org/abs/2511.23443}
}

@inproceedings{zhou2024pretrained,
  title={Pre-trained Large Language Models Use Fourier Features to Compute Addition},
  author={Tianyi Zhou and Vatsal Sharan and Robin Jia},
  booktitle={The Thirty-eighth Annual Conference on Neural Information Processing Systems},
  year={2024},
  url={https://arxiv.org/abs/2406.03445}
}

@inproceedings{fernandez2025sinusoidal,
  title={Sinusoidal Initialization, Time for a New Start},
  author={Fern{\'a}ndez-Hern{\'a}ndez, Alberto and
          Mestre, Jose I. and
          Dolz, Manuel F. and
          Duato, Jose and
          Quintana-Ort{\'i}, Enrique S.},
  booktitle={The Thirty-ninth Annual Conference on Neural Information Processing Systems},
  year={2025},
  url={https://openreview.net/forum?id=FGliQVcrDZ}
}

@inproceedings{gillman2025fourier,
  title={Fourier Head: Helping Large Language Models Learn Complex Probability Distributions},
  author={Nate Gillman and Daksh Aggarwal and Michael Freeman and Saurabh Singh and Chen Sun},
  booktitle={The Thirteenth International Conference on Learning Representations},
  year={2025},
  url={https://arxiv.org/abs/2410.22269}
}

@article{sun2025penguin,
  title={PENGUIN: Enhancing Transformer with Periodic-Nested Group Attention for Long-term Time Series Forecasting},
  author={Tian Sun and Yuqi Chen and Weiwei Sun},
  journal={arXiv preprint arXiv:2508.13773},
  year={2025},
  url={https://arxiv.org/abs/2508.13773}
}

@article{sutton2019bitter,
  title={The bitter lesson},
  author={Sutton, Richard S},
  journal={Incomplete Ideas (blog)},
  volume={13},
  number={1},
  pages={38},
  year={2019},
  url={http://www.incompleteideas.net/IncIdeas/BitterLesson.html}
}

@inproceedings{
prieto2025grokking,
title={Grokking at the Edge of Numerical Stability},
author={Lucas Prieto and Melih Barsbey and Pedro A. M. Mediano and Tolga Birdal},
booktitle={The Thirteenth International Conference on Learning Representations},
year={2025},
url={https://openreview.net/forum?id=TvfkSyHZRA}
}

@inproceedings{abramov2025grokking,
  title={Grokking in the Wild: Data Augmentation for Real-World Multi-Hop Reasoning with Transformers},
  author={Abramov, Roman and Steinbauer, Felix and Kasneci, Gjergji},
  booktitle={Proceedings of ICML},
  year={2025}
}

@inproceedings{
thilak2022slingshot,
title={The Slingshot Mechanism: An Empirical Study of Adaptive Optimizers and the {\textbackslash}emph\{Grokking Phenomenon\}},
author={Vimal Thilak and Etai Littwin and Shuangfei Zhai and Omid Saremi and Roni Paiss and Joshua M. Susskind},
booktitle={Has it Trained Yet? NeurIPS 2022 Workshop},
year={2022},
url={https://openreview.net/forum?id=lY1e0PNkSJ}
}

@inproceedings{power2022grokking,
  title={Grokking: Generalization beyond overfitting on small algorithmic datasets},
  author={Power, Alethea and Burda, Yuri and Edwards, Harri and Babuschkin, Igor and Misra, Vedant},
  booktitle={ICLR Workshop on Mathematical Reasoning},
  year={2022}
}

@inproceedings{
chughtai2023neural,
title={Neural Networks Learn Representation Theory: Reverse Engineering how Networks Perform Group Operations},
author={Bilal Chughtai and Lawrence Chan and Neel Nanda},
booktitle={ICLR 2023 Workshop on Physics for Machine Learning},
year={2023},
url={https://openreview.net/forum?id=j4_YHiTAN63}
}

@inproceedings{zhong2023clock,
  title={The Clock and the Pizza: Two Stories in Mechanistic Explanation of Neural Networks},
  author={Zhong, Ziqian and Liu, Ziming and Tegmark, Max and Andreas, Jacob},
  booktitle={Advances in Neural Information Processing Systems (NeurIPS)},
  year={2023}
}

@inproceedings{
morwani2024feature,
title={Feature emergence via margin maximization: case studies in algebraic tasks},
author={Depen Morwani and Benjamin L. Edelman and Costin-Andrei Oncescu and Rosie Zhao and Sham M. Kakade},
booktitle={The Twelfth International Conference on Learning Representations},
year={2024},
url={https://openreview.net/forum?id=i9wDX850jR}
}

@article{lee2024grokfast,
  title={Grokfast: Accelerated Grokking by Amplifying Slow Gradients},
  author={Lee, Jaerin and Kang, Bong Gyun and Kim, Kihoon and Lee, Kyoung Mu},
  journal={arXiv preprint arXiv:2405.20233},
  year={2024}
}

@article{truong2026norm,
  title={The Norm-Separation Delay Law of Grokking: A First-Principles Theory of Delayed Generalization},
  author={Truong, Xuan Khanh and Truong, Quynh Hoa and Luu, Duc Trung and Phan, Thanh Duc},
  journal={arXiv preprint arXiv:2603.13331},
  year={2026}
}

@article{musat2026geometry,
  title={The Geometry of Grokking: Norm Minimization on the Zero-Loss Manifold},
  author={Musat, Tiberiu},
  journal={arXiv preprint arXiv:2511.01938},
  year={2026}
}

@misc{loshchilov2025ngptnormalized,
      title={nGPT: Normalized Transformer with Representation Learning on the Hypersphere}, 
      author={Ilya Loshchilov and Cheng-Ping Hsieh and Simeng Sun and Boris Ginsburg},
      year={2025},
      eprint={2410.01131},
      archivePrefix={arXiv},
      primaryClass={cs.LG},
      url={https://arxiv.org/abs/2410.01131}, 
}

\appendix
\section{Experimental Details and Hyperparameters}
\label{app:hyperparameters}

All experiments were implemented in PyTorch and executed deterministically across 10 random seeds to ensure exact reproducibility. Both tasks utilized the same core Transformer architecture, with task-specific optimization parameters as outlined below.

\subsection{Core Architecture}
\begin{itemize}
\item \textbf{Embedding / Residual Dimension ($d_{\text{model}}$):} 128
\item \textbf{Attention Heads:} 4
\item \textbf{Head Dimension ($d_{\text{head}}$):} 32
\item \textbf{MLP Hidden Dimension ($d_{\text{mlp}}$):} 512
\item \textbf{Activation Function:} ReLU
\item \textbf{Layers:} 1
\end{itemize}

\subsection{Task 1: Modular Addition ($\mathbb{Z}_{113}$)}
\begin{itemize}
\item \textbf{Sequence Format:} \texttt{[token\_a, token\_b, equals\_token]}
\item \textbf{Vocabulary Size:} 114 (tokens 0--112, plus 113 as the operator token)
\item \textbf{Dataset Split:} $30\%$ Train ($\sim 3,830$ samples) / $70\%$ Test ($\sim 8,939$ samples)
\item \textbf{Optimizer:} AdamW ($\beta_1=0.9, \beta_2=0.999$). For the Uniform Attention Ablation experiments (Intervention B), $\beta_2$ was set to $0.98$.
\item \textbf{Batch Size:} Full-batch (all training samples processed simultaneously)
\item \textbf{Learning Rate:} $1 \times 10^{-4}$ and $6 \times 10^{-4}$
\item \textbf{Weight Decay:} $1.0$ (Baselines \& Sphere $\lambda=1.0$) / $0.0$ (Bounded Sphere $\lambda=0.0$)
\item \textbf{Max Epochs:} 15,000 to 100,000 (depending on configuration)
\end{itemize}

\subsection{Task 2: Permutation Composition ($S_5$)}
\begin{itemize}
\item \textbf{Sequence Format:} \texttt{[token\_a, token\_b, equals\_token]}
\item \textbf{Vocabulary Size:} 121 (tokens 0--119, plus 120 as the operator token)
\item \textbf{Dataset Split:} $30\%$ Train ($4,320$ samples) / $70\%$ Test ($10,080$ samples)
\item \textbf{Optimizer:} AdamW ($\beta_1=0.9, \beta_2=0.999$)
\item \textbf{Batch Size:} Full-batch
\item \textbf{Learning Rate:} $1 \times 10^{-3}$
\item \textbf{Weight Decay:} $1.0$ (Baselines \& Sphere $\lambda=1.0$) / $0.0$ (Bounded Sphere $\lambda=0.0$)
\item \textbf{Max Epochs:} 100,000
\end{itemize}

\section{Uniform Attention Ablation on $S_5$}
\label{app:s5_cbow}
 
To complete the factorial evaluation of our two interventions across both tasks, we applied the Uniform Attention Ablation (Intervention~B) to the $S_5$ permutation composition task. Models were trained for 100,000 epochs with a learning rate of $10^{-3}$ across 5 independent seeds. Results are summarized in Table~\ref{tab:s5_cbow}.
 
\begin{table}[h]
\centering
\caption{Peak test accuracy under Uniform Attention Ablation on $S_5$ permutation composition across 5 random seeds.}
\label{tab:s5_cbow}
\small
\begin{tabular}{lccc}
\toprule
Architecture (Uniform Attention) & Mean Peak Acc. & Max Peak Acc. & Success Rate (100\%) \\
\midrule
LayerNorm Baseline (WD=1.0) & 99.12\% & 99.42\% & 0\,/\,5 \\
RMSNorm Baseline (WD=1.0) & 99.65\% & 99.94\% & 0\,/\,5 \\
Bounded Sphere ($\lambda = 1.0$) & 2.84\% & 3.42\% & 0\,/\,5 \\
Bounded Sphere ($\lambda = 0.0$) & 2.41\% & 2.54\% & 0\,/\,5 \\
\bottomrule
\end{tabular}
\end{table}
 
The results reveal two findings. First, the spherical constraint remains incompatible with $S_5$ regardless of attention type---both bounded configurations fail to exceed chance-level accuracy ($\sim$2--3\% for 120 classes), confirming that the topology, not the routing mechanism, is the primary barrier. Second, while LayerNorm and RMSNorm baselines achieve near-perfect peak accuracy under uniform attention, they do not exhibit meaningful acceleration compared to the learned-attention setting (Section~4.4), indicating that data-dependent routing is not the bottleneck for this non-commutative task.

\section{Relationship to nGPT}
\label{app:ngpt}

Because both our Intervention~A and nGPT place representations on a unit hypersphere, we provide a detailed comparison to clarify that the two works differ in goal, design philosophy, and every architectural detail beyond the shared use of $L_2$ normalization on the hidden state.

\subsection{Different Goals}

nGPT is an \emph{architecture proposal}: it seeks to build a faster Transformer for general-purpose language modeling and reports $4$--$20\times$ training speedups on OpenWebText. Its success criterion is downstream task performance.

Our bounded spherical topology is a \emph{mechanistic probe}: it tests whether removing the radial degree of freedom from the residual stream changes the memorization-to-generalization transition on tasks whose circuit-level solutions are already characterized. Its success criterion is not faster training per se, but the \emph{differential outcome} between tasks---acceleration on $\mathbb{Z}_p$ and failure on $S_5$---which constitutes evidence about the relationship between architectural geometry and task symmetry. A probe that always succeeds would be uninformative; the failure mode is half the contribution.

\subsection{Architectural Differences}

Despite superficial similarity, every design decision diverges.

\paragraph{Scope of normalization.}
nGPT normalizes \emph{all weight matrices} ($W_q$, $W_k$, $W_v$, $W_o$, $W_u$, $W_\nu$, $W_o^{\mathrm{MLP}}$, $E_{\mathrm{input}}$, $E_{\mathrm{output}}$) along their embedding dimension, converting all matrix-vector products into cosine similarities. Our intervention normalizes only the \emph{residual stream activations} (the hidden state $\mathbf{h}$). We do not constrain any weight matrix.

\paragraph{Update rule.}
nGPT replaces standard residual addition with an interpolation controlled by learnable per-dimension eigen learning rates:
\begin{equation}
    \mathbf{h} \leftarrow \mathrm{Norm}\!\left(\mathbf{h} + \boldsymbol{\alpha}_T\!\left(\mathrm{Norm}(\mathbf{h}_T) - \mathbf{h}\right)\right),
\end{equation}
where $\boldsymbol{\alpha}_T \in \mathbb{R}_{\geq 0}^{d_{\mathrm{model}}}$ is a learned vector that modulates how far the hidden state moves toward each block's output. Our update is standard residual addition followed by hard projection:
\begin{equation}
    \mathbf{h} \leftarrow \Pi_S\!\left(\mathbf{h} + \mathrm{Block}(\mathbf{h})\right),
\end{equation}
where $\Pi_S$ is a parameter-free $L_2$ normalization. There is no interpolation and no learnable step size.

\paragraph{Compensatory scaling.}
nGPT introduces six families of learnable scaling parameters ($\mathbf{s}_{qk}$, $\mathbf{s}_u$, $\mathbf{s}_\nu$, $\mathbf{s}_z$, $\boldsymbol{\alpha}_A$, $\boldsymbol{\alpha}_M$) that allow the network to locally modulate or effectively bypass the geometric constraint along individual dimensions. These are essential to nGPT's success on heterogeneous language data: without them, the strict spherical constraint would be too restrictive for tasks whose solutions do not align with spherical geometry.

Our design is \emph{deliberately uncompensated}. We provide no learnable escape mechanism. This rigidity is not a limitation---it is the experimental design. A probe with compensatory flexibility would confound the question we are asking: \emph{does the radial degree of freedom influence grokking dynamics?} If the network could learn to circumvent the constraint, a positive result would be uninterpretable.

\paragraph{Normalization layers.}
nGPT removes all LayerNorm/RMSNorm layers entirely, replacing them with its interpolation-plus-projection scheme. In our baseline comparisons, we evaluate models \emph{with} LayerNorm and RMSNorm as controls. Our spherical projection replaces LayerNorm only in the bounded configurations.

\paragraph{Weight decay.}
nGPT eliminates weight decay, since all weight norms are fixed by construction. We explicitly vary weight decay ($\lambda = 0.0$ vs.\ $\lambda = 1.0$) as an experimental variable to study its interaction with the bounded topology---finding that weight decay introduces optimization friction under the spherical constraint (Section~\ref{sec:results}).

\paragraph{Output topology.}
Both approaches normalize the unembedding matrix, but nGPT uses a \emph{learnable per-token} scaling vector $\mathbf{s}_z \in \mathbb{R}^V$, while we use a single \emph{fixed} temperature scalar $\tau = 10.0$. Again, ours is deliberately inflexible.

\subsection{Summary of Differences}

\begin{table}[h]
\centering
\caption{Architectural comparison between nGPT and our Intervention~A.}
\label{tab:ngpt-comparison}
\small
\begin{tabular}{lll}
\toprule
\textbf{Design Choice} & \textbf{nGPT} & \textbf{Ours (Intervention A)} \\
\midrule
Goal & Faster LM pretraining & Mechanistic hypothesis test \\
Weight matrix normalization & All matrices, all layers & None \\
Hidden state update & Learnable interpolation & Residual add + hard projection \\
Learnable step sizes & Per-dimension $\boldsymbol{\alpha}$ & None \\
Compensatory scaling & 6 parameter families & None \\
Unembedding scaling & Learnable per-token $\mathbf{s}_z$ & Fixed scalar $\tau$ \\
Weight decay & Removed & Experimentally varied \\
Normalization layers & Removed & Replaced / retained (baselines) \\
Negative control & N/A & $S_5$ permutation composition \\
\bottomrule
\end{tabular}
\end{table}

\subsection{Why the Distinction Matters}

The key difference is not a list of architectural details but the role that \emph{rigidity} plays in each design.

nGPT's compensatory parameters allow it to succeed on language modeling---a domain containing heterogeneous subtasks with diverse symmetry structures. This flexibility is architecturally necessary but mechanistically opaque: when nGPT achieves a $10\times$ speedup, it is unclear which subtasks benefit from the spherical geometry and which are rescued by the learned scaling factors.

Our rigid constraint answers a narrower but sharper question. On modular arithmetic, the constraint eliminates the memorization phase entirely---demonstrating that the radial degree of freedom is sufficient to explain the grokking delay in this setting. On $S_5$, the same constraint prevents generalization completely---demonstrating that the acceleration is not a generic regularization effect but depends on alignment between the constraint and the task's intrinsic symmetry.

Together, these results provide a framework for understanding \emph{when and why} hypersphere constraints accelerate learning. nGPT's success on language suggests that natural language contains sufficient periodic or angular structure to benefit from spherical geometry, but its compensatory mechanisms indicate that not all subtasks are so aligned. Our $S_5$ result makes this tension precise: even the simplest non-commutative algebraic task resists a rigid spherical constraint, implying that nGPT's per-dimension scaling factors are not merely convenient but \emph{structurally necessary} for accommodating non-cyclic structure within a globally spherical architecture.

\end{document}